\documentclass[pdflatex,sn-nature]{sn-jnl}

\usepackage{graphicx}%
\usepackage{multirow}%
\usepackage{amsmath,amssymb,amsfonts}%
\usepackage{amsthm}%
\usepackage{mathrsfs}%
\usepackage[title]{appendix}%
\usepackage{xcolor}%
\usepackage{textcomp}%
\usepackage{manyfoot}%
\usepackage{booktabs}%
\usepackage{algorithm}%
\usepackage{algorithmicx}%
\usepackage{algpseudocode}%
\usepackage{listings}%

\theoremstyle{thmstyleone}%
\theoremstyle{thmstyletwo}%

\theoremstyle{thmstylethree}%

\begin{document}

\title[Article Title]{End to End AI System for Surgical Gesture Sequence Recognition and Clinical Outcome Prediction}


\author[1]{\fnm{Xi} \sur{Li}}\email{xi.li2@cshs.org}

\author[1]{\fnm{Nicholas} \sur{Matsumoto}}\email{nicholas.matsumoto@cshs.org}

\author[2]{\fnm{Ujjwal} \sur{Pasupulety}}\email{ujjwal.pasupulety@cshs.org}

\author[2]{\fnm{Atharva} \sur{Deo}}\email{atharva.deo@cshs.org}

\author[2]{\fnm{Cherine} \sur{Yang}}\email{cherine.yang@cshs.org}

\author[1]{\fnm{Jay} \sur{Moran}}\email{jay.moran@cshs.org}

\author[1]{\fnm{Miguel E.} \sur{Hernandez}}\email{miguel.e.hernandez@cshs.org}

\author[2]{\fnm{Peter} \sur{Wager}}\email{peter.wager@cshs.org}

\author[2]{\fnm{Jasmine} \sur{Lin}}\email{jasmine.lin@cshs.org}

\author[2]{\fnm{Jeanine} \sur{Kim}}\email{sun.kim@cshs.org}

\author[3]{\fnm{Alvin C.} \sur{Goh}}\email{goha@mskcc.org}

\author[4]{\fnm{Christian} \sur{Wagner}}\email{wagner@st-antonius-gronau.de}

\author[5]{\fnm{Geoffrey A.} \sur{Sonn}}\email{gsonn@stanford.edu}

\author*[2, 6]{\fnm{Andrew J.} \sur{Hung}}\email{andrew.hung@cshs.org}

\affil[1]{\orgdiv{Department of Computational Biomedicine, Center for Artificial Intelligence Research and Education}, \orgname{Cedars Sinai Medical Center}, \orgaddress{\city{Los Angeles}, \state{CA}, \country{USA}}}

\affil*[2]{\orgdiv{Department of Urology}, \orgname{Cedars Sinai Medical Center}, \orgaddress{\city{Los Angeles}, \state{CA}, \country{USA}}}

\affil[3]{\orgdiv{Department of Urology}, \orgname{Memorial Sloan Kettering Cancer Center}, \orgaddress{\city{New York}, \state{NY}, \country{USA}}}

\affil[4]{\orgdiv{Department of Urology, Pediatric Urology and Urologic Oncology}, \orgname{St. Antonius-Hospital}, \orgaddress{\city{Gronau}, \country{Germany}}}

\affil[5]{\orgdiv{Department of Urology}, \orgname{Stanford University Medical Center}, \orgaddress{\city{Stanford}, \state{CA}, \country{USA}}}

\affil[6]{\orgdiv{Department of Computational Biomedicine}, \orgname{Cedars Sinai Medical Center}, \orgaddress{\city{Los Angeles}, \state{CA}, \country{USA}}\vspace{2em}}

\affil[]{\small\textbf{Proprietary Notice:} This end-to-end AI system for surgical gesture sequence recognition and clinical outcome prediction, known as Frame-to-Outcome (F2O), is owned by and proprietary to Cedars-Sinai Medical Center. © 2025 Cedars-Sinai Medical Center. All rights reserved.
For any use requests, please reach out to CSTechTransfer@cshs.org}

\abstract{Fine-grained analysis of intraoperative behavior and its impact on patient outcomes remain a longstanding challenge. We present Frame-to-Outcome (F2O), an end-to-end system that translates tissue dissection videos into gesture sequences and uncovers patterns associated with postoperative outcomes. Leveraging transformer-based spatial and temporal modeling and frame-wise classification, F2O robustly detects consecutive short (\~2 seconds) gestures in the nerve-sparing step of robot-assisted radical prostatectomy (AUC: 0.80 frame-level; 0.81 video-level). F2O-derived features—gesture frequency, duration, and transitions—predicted postoperative outcomes with accuracy comparable to human annotations (0.79 vs. 0.75; overlapping 95\% CI). Across 25 shared features, effect size directions were concordant with small differences (\(\Delta d_{\text{avg}} \approx 0.07\)), and strong correlation (\(r = 0.96 \), \( p < 1\times10^{-14}\)). F2O also captured key patterns linked to erectile function recovery, including prolonged tissue peeling and reduced energy use. By enabling automatic interpretable assessment, F2O establishes a foundation for data-driven surgical feedback and prospective clinical decision support.}




\maketitle

\section{Introduction}\label{sec1}

The automated quantification of intraoperative surgical activity and analysis of its relationship to clinical outcomes remain a fundamental challenge in surgical data science \cite{levendovics2024surgical, maier2022surgical,knudsen2024clinical}. While high-resolution robotic and endoscopic video recordings have become more common in modern operating rooms \cite{kaan2020clinical}, the majority of procedures still lack standardized, automated tools for capturing and interpreting intraoperative activities. This gap presents a significant barrier to advancing evidence-based surgery, where novel data-driven insights into performance are critical \cite{loukas2018video}. 

Several longstanding challenges have limited progress in this area. One is the inconsistent and non-standardized surgical terminology \cite{van2021gesture} , which has led researchers to develop alternative dictionaries and taxonomies tailored to specific procedures or datasets \cite{gao2014jhu,ma2021novel,ye2025comprehensive}. Another major bottleneck is the labor-intensive nature of manual video annotation. It requires domain expertise, is highly time-consuming, and is prone to inter-rater variability and subjective interpretation \cite{maier2022surgical,van2021gesture,ward2021challenges}. In addition, the high variability in technique across patients, surgeons, procedures, and institutions further complicates model development. Factors such as anatomical differences, the presence of blood, camera motion, and variations in surgical style introduce considerable heterogeneity into video data \cite{van2021gesture}. Furthermore, the technical difficulty of modeling fine-grained surgical actions poses significant obstacles to automated recognition \cite{van2021gesture}. Collectively, these barriers limit access to granular standardized intraoperative surgical data, constrain assessments of surgical technique, and restrict investigation into relationships between surgical performance and patient outcomes.

\begin{figure}[htbp]
\centering
\includegraphics[width=\textwidth]{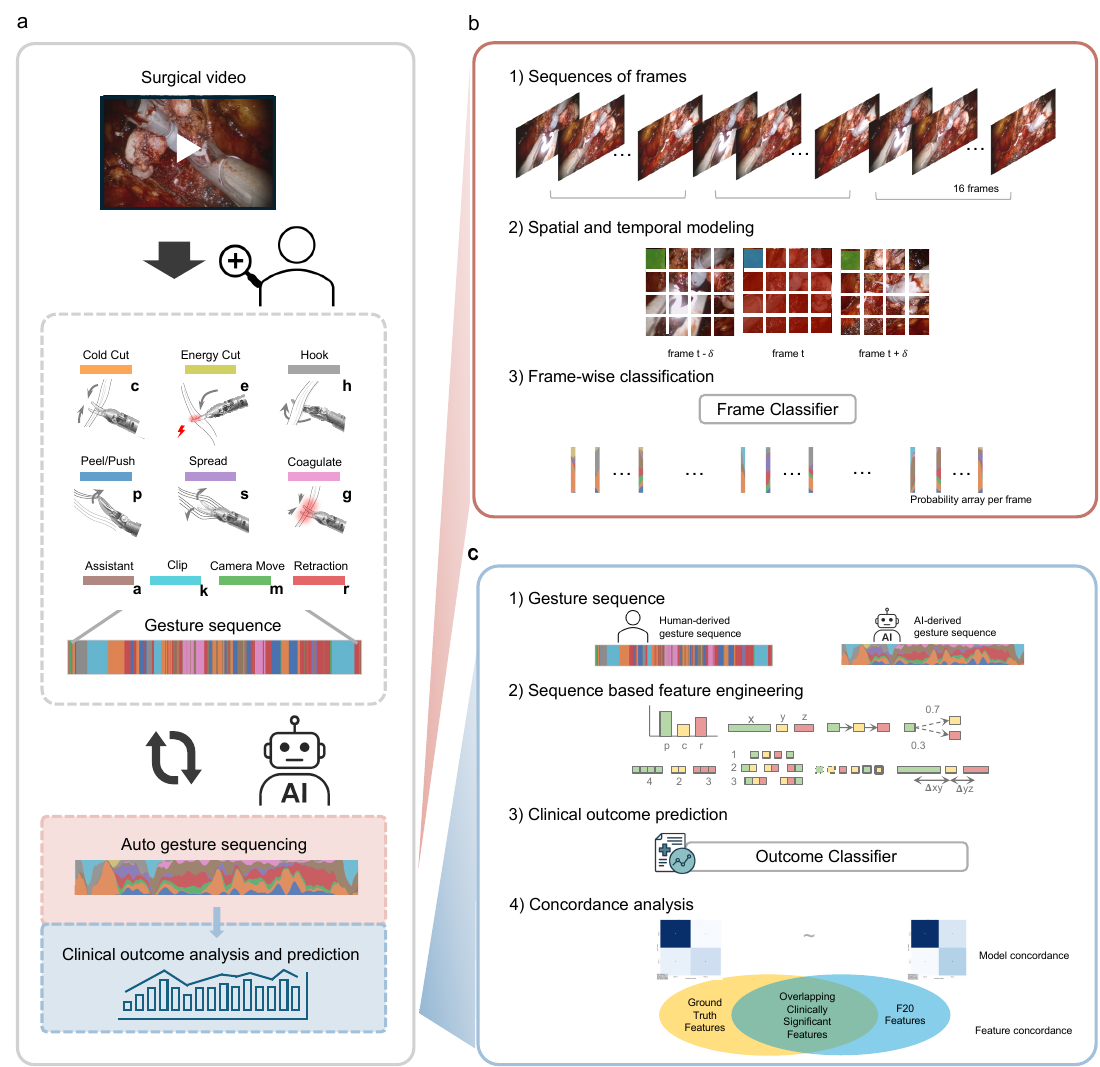}
\caption{Frame-to-Outcome (F2O) is an end-to-end AI system for surgical gesture sequence recognition and clinical outcome analysis. \textbf{a}, Surgical videos, such as those from the nerve-sparing (NS) step of robot-assisted radical prostatectomy (RARP), are annotated by trained human raters who identify over ten dominant gesture classes and label the start and end times of each gesture. Each video typically contains \~270 gestures over a 10-minute duration, with an average gesture lasting 2 seconds. Frame-to-Outcome (F2O) automates the recognition of these fine-grained gestures and enables downstream clinical outcome analysis. \textbf{b}, The system processes untrimmed tissue dissection videos and outputs a sequence of standardized surgical gesture probabilities by combining spatial and temporal modeling with frame-wise classification. Specifically, it processes sequences of 16 frames, leveraging spatial and temporal neighbors (red and green) to compute self-attention for each target patch (blue). These context-aware embeddings are then passed through a frame-wise classifier, which produces gesture probability distributions for each frame based on the aggregated representations. \textbf{c}, Sequence-based feature engineering is then applied to identify relationships with clinical outcomes, and results are evaluated through concordance analysis, including both feature-level and model-level concordance.
}\label{Fig. 1}
\end{figure}

Surgical gestures, defined as the smallest meaningful unit of surgical instrument and human tissue interaction, capture the intentional and discrete actions performed by a surgeon during a procedure. Each gesture reflects a specific movement shaped by anatomical context and surgical intent, such as “spreading tissue” or “cutting tissue”. When combined, these sequences of gestures form complex procedures and offer an interpretable abstraction for modeling skill, workflow, and decision-making \cite{ma2025surgical}. In robot-assisted radical prostatectomy (RARP), for example, the nerve-sparing (NS) step consists of approximately 270 such gestures within a 10-minute segment, each lasting an average of 2 seconds (Fig. \ref{Fig. 1}a). Accurate identification and classification of these gestures provide a critical foundation for large-scale, data-driven surgical analytics. Our previous studies introduced a gesture classification system to standardize surgical terminology and consistently measure surgical gestures \cite{ma2021novel}. Additional efforts have also identified the relationship between the sequences of standardized surgical gestures and postoperative outcomes, establishing surgical gestures’ role as fine-grained metrics for surgical quality and patient prognostication \cite{ma2022surgical}.

Recent advances in deep learning, particularly transformer-based architectures, have demonstrated remarkable success in capturing spatiotemporal patterns across various domains, including human activity recognition \cite{vahdani2022deep}, temporal action localization \cite{9062498}, action segmentation \cite{ding2023temporal}. In the surgical data science area \cite{min2025innovating}, prior efforts achieved per-clipped gesture classification with human-annotated start and end times \cite{kiyasseh2023vision}. However, applying these models to surgical gesture recognition within its natural sequence of gestures poses unique challenges, given the fine temporal resolution, rapid transitions, and high inter-case variability. Some approaches have leveraged kinematic signals to approximate transitions \cite{van2022gesture}, while others have applied fixed time-interval inference followed by post-hoc aggregation to construct gesture sequences \cite{kiyasseh2023vision}. Yet, these methods often require diverse and extensive annotations or fall short in capturing the appropriate duration of gestures.

To address these challenges, we developed Frame-to-Outcome (F2O), an end-to-end system that automatically recognizes fine-grained gestures in surgical video and supports downstream clinical outcome analysis. Leveraging Transformer-based spatial and temporal modeling and frame-wise classification, along with standardized surgical gesture terminology and sequence feature engineering, the system translates untrimmed tissue dissection videos to sequences of consecutive standardized surgical gestures without human efforts and identifies important gesture patterns associated with the outcome. By utilizing 294 annotated RARP NS videos from 4 international centers, spanning 23 surgeons and 10 dominant gesture classes (Fig. \ref{Fig. 1}a), the model demonstrated strong performance at both frame-level and video-level. Its predictions aligned closely with ground-truth features in downstream outcome analysis, confirming both technical accuracy and clinical relevance. Additional experiments across backbones and data scales showed consistent performance, highlighting the system’s adaptability, efficiency, and suitability for real-world clinical deployment. Together, this work lays the groundwork for real-time intraoperative feedback, automated video annotation, and large-scale surgical analytics that link technical performance with patient outcomes.

\section{Results}\label{sec2}

\subsection{F2O achieves accurate frame-level classification across surgical gesture classes}\label{subsec2}

The core of the F2O pipeline is built on precise frame-level gesture classification. To model the fine temporal granularity of surgical gestures (Fig. \ref{Fig. 1}a), the system integrates a Transformer-based spatial and temporal modeling \cite{bertasius2021space} with a frame-wise classifier \cite{chen2022frame} (Fig. \ref{Fig. 1}b). This architecture is enhanced by stratified and adaptive overlapping sequence sampling, which prioritizes underrepresented gestures based on their prevalence, and by targeted data augmentation strategies. 

\begin{figure}[htbp]
\centering
\includegraphics[width=\textwidth]{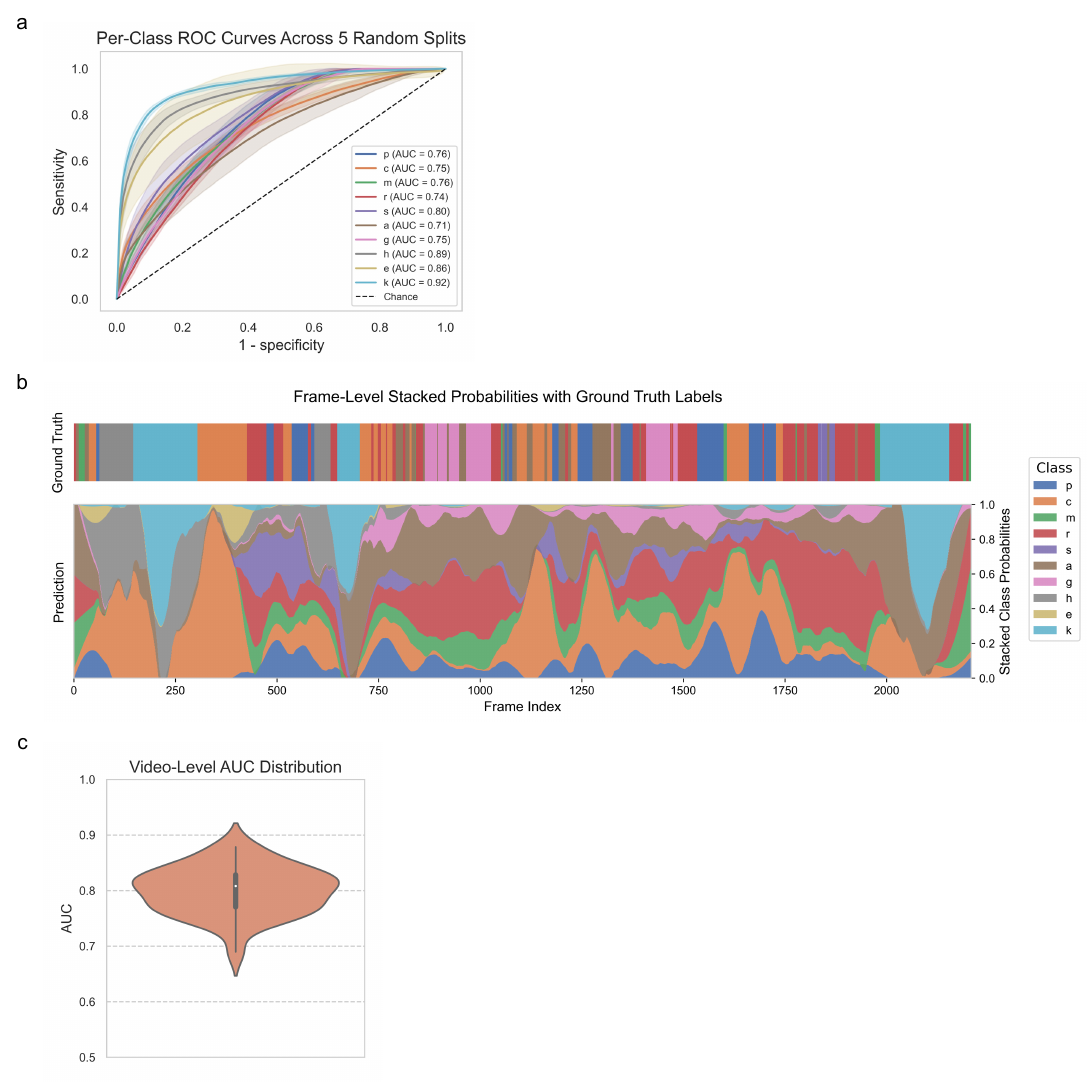}
\caption{F2O accurately classifies frames across classes and videos. \textbf{a}, Frame-level performance across gesture classes of \textit{cold cut} (c), \textit{hook} (h), \textit{clip} (k), \textit{camera move} (m), \textit{peel} (p), \textit{retraction} (r), \textit{spread} (s), \textit{assistant} (a), \textit{coagulation} (g), and \textit{energy cut} (e), evaluated over five randomized data splits. \textbf{b}, Temporal alignment between model-predicted gesture probabilities and ground truth when applied frame-by-frame to the entire video. \textbf{c}, Video-level performance across 29 full-length videos in the test set, showing the distribution of AUC values.
}\label{Fig. 2}
\end{figure}

To evaluate robustness, we conducted five random data splits and measured the area under the receiver operating characteristic curve (AUC). The model achieved 0.80 (95\% CI: 0.78–0.81) on the test set. Per-class ROC curves from the 5 test splits are shown (Fig. \ref{Fig. 2}a). The model demonstrated consistent classification performance across ten dominant gesture classes, which cover approximately 97\% of gestures in a video. Notably, the gesture \textit{clip} (k) achieved the highest performance (AUC = 0.92), followed by \textit{hook} (h) (AUC = 0.89), and \textit{energy cut} (e) (AUC = 0.86), while the remaining seven gesture classes yielded AUCs ranging from 0.71 to 0.80. These findings demonstrate the capability of F2O to learn temporally precise features that are necessary for fine-grained gestures across various class types.

\subsection{F2O enables automatic gesture sequence recognition across videos }\label{subsec2}

To simulate real-world clinical settings, we applied the trained F2O system to 29 full-length videos in the test dataset, each capturing continuous tissue dissection activity during the NS section of RARP. The model predicted gestures continuously across entire recordings. Despite the pronounced class imbalance characteristic of surgical procedures, where some gestures are frequent and others are rare, the model maintained consistently strong performance across all test videos.

To illustrate the temporal coherence of gesture predictions, Fig. \ref{Fig. 2}b presents stacked class probability plots for a representative video. The model’s predictions exhibit smooth and interpretable transitions over time, with pronounced shifts in the predicted class distribution closely aligned with ground-truth gesture annotations (indicated in the top bar). For instance, the probability for the \textit{clip} gesture (Class k) remains close to zero for most of the video, except for three distinct peaks that correspond precisely to the \textit{clip} events in the ground truth. This alignment underscores the model’s capacity for fine-grained temporal localization, even for short-duration gestures within complex procedures.

Fig. \ref{Fig. 2}c, shows the distribution of video-level AUCs across the 29 videos, ranging mostly from 0.77 to 0.83, with a median AUC near 0.81. The distribution was moderately symmetrical, with only a few outlier videos exhibiting below 0.70. These results demonstrate the model’s robustness in handling continuous, heterogeneous surgical recordings and its capacity to generalize effectively beyond the training examples. Collectively, these results demonstrate that the F2O pipeline not only achieves accurate frame-level classification but also enables automatic recognition of complete gesture sequences in surgical videos. This capability offers strong potential for real-time surgical analysis, retrospective review, and scalable annotation of large-scale surgical datasets.

\subsection{Predicting clinical outcomes using engineered features from surgical gesture sequences}\label{subsec3}

To assess the clinical utility of automatically predicted surgical gestures, we evaluated whether sequence based features engineered from F2O-predicted gesture sequences could accurately predict postoperative erectile function (EF) recovery at 12 months. We compared these predictions to those generated using features derived from manually annotated ground-truth gesture sequences. All predictions were performed using a tabular transformer model trained to classify binary EF outcome based on engineered gesture features (see Methods).

\begin{figure}[htbp]
    \centering
    \includegraphics[width=1\linewidth]{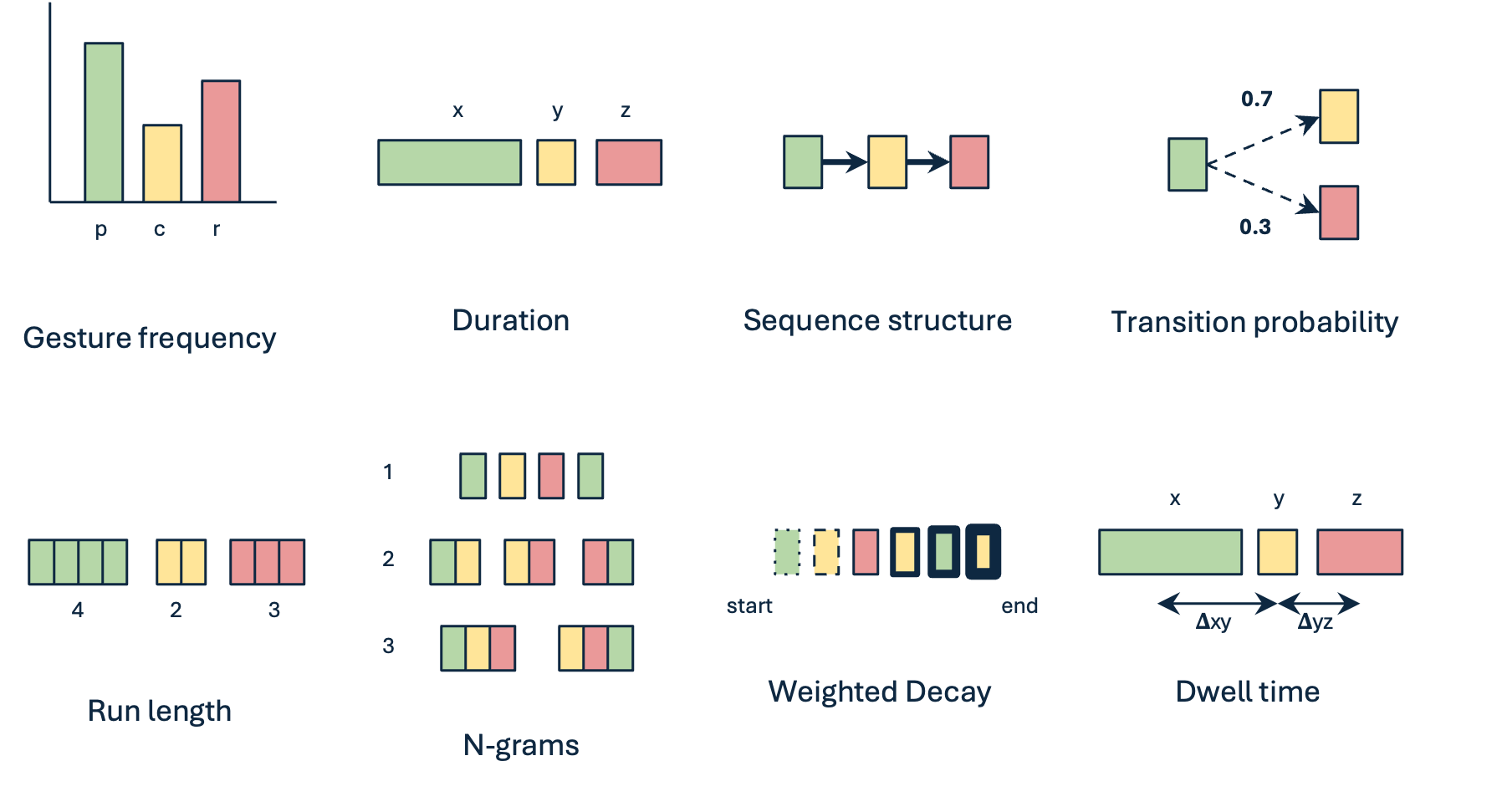}
    \caption{Sequence based features}
    \label{fig:gesture_desc}
\end{figure}

Engineered features included gesture frequency, transition entropy, gesture duration statistics, and temporal sequence patterns Fig. \ref{fig:gesture_desc}. The same feature engineering pipeline was applied to both ground truth and F2O-derived gesture sequences. Models were trained using stratified 5-fold cross-validation and evaluated by accuracy with 95\% confidence intervals computed across splits.

The tabular transformer trained on F2O-derived features to determine a binary classification of EF outcome achieved a mean accuracy of 0.7902 (95\% CI: [0.6696, 0.9109]). In comparison, the model trained on human annotated ground-truth gesture features achieved a mean accuracy of 0.7506 (95\% CI: [0.6397, 0.8616]). Performance across both input sources was closely aligned, with overlapping confidence intervals and similar classification metrics, suggesting strong concordance between predicted and manually annotated gesture-derived representations. 

Interestingly, the mean accuracy of the model trained on F2O-derived features exceeded that of the ground-truth-based model, despite being fully automated. This result suggests that F2O may not only replicate the clinically relevant information embedded in manual annotations, but in some cases improve upon it by producing more temporally consistent or less noisy representations. These findings reinforce the utility of F2O as a scalable and potentially more robust alternative to manual gesture annotation in surgical outcome modeling where human annotation may introduce bias and inconsistency.

These results indicate that F2O-predicted gesture sequences preserve clinically relevant structure and support downstream clinical outcome analysis as its comparable model performance underscores the feasibility of using F2O for large-scale, automated surgical outcome studies without requiring exhaustive manual labeling.

\subsection{Concordance of clinically informative gesture features}\label{subsec4}
To further assess the fidelity of the gesture sequences predicted by F2O, we compared the clinical significance of engineered features derived from F2O and manually annotated sequences. Specifically, we evaluated whether both sources identify similar gesture metrics as being associated with erectile function recovery at 12 months.

We conducted two-tailed Student’s t-tests (N = 138) for each of 2,484 candidate gesture-derived features, using the binary clinical outcome as the target. Features were then ranked by their p-values in both the F2O-derived and ground-truth sets. Among the top 50 features from each source, 25 overlapped (50\%), indicating that F20 captures half of the most clinically relevant features detectable through manual annotation. This overlap suggests strong agreement in outcome-relevant signal, despite the absence of human intervention in the F2O pipeline.

As the output from the frame level classification is not in the format of the human annotated data, where it is a sequence of gestures with timestamps, the frame classifications are reformatted using a change-point detection (see Methods). Using the optimized change‐point detection parameters and gesture‐specific weights, F2O produced a sequence of gestures which subsequently go through feature engineering from which we can identify  most strongly correlated with clinical outcome. 

\begin{figure}[htbp]
    \centering
    \includegraphics[width=1\linewidth]{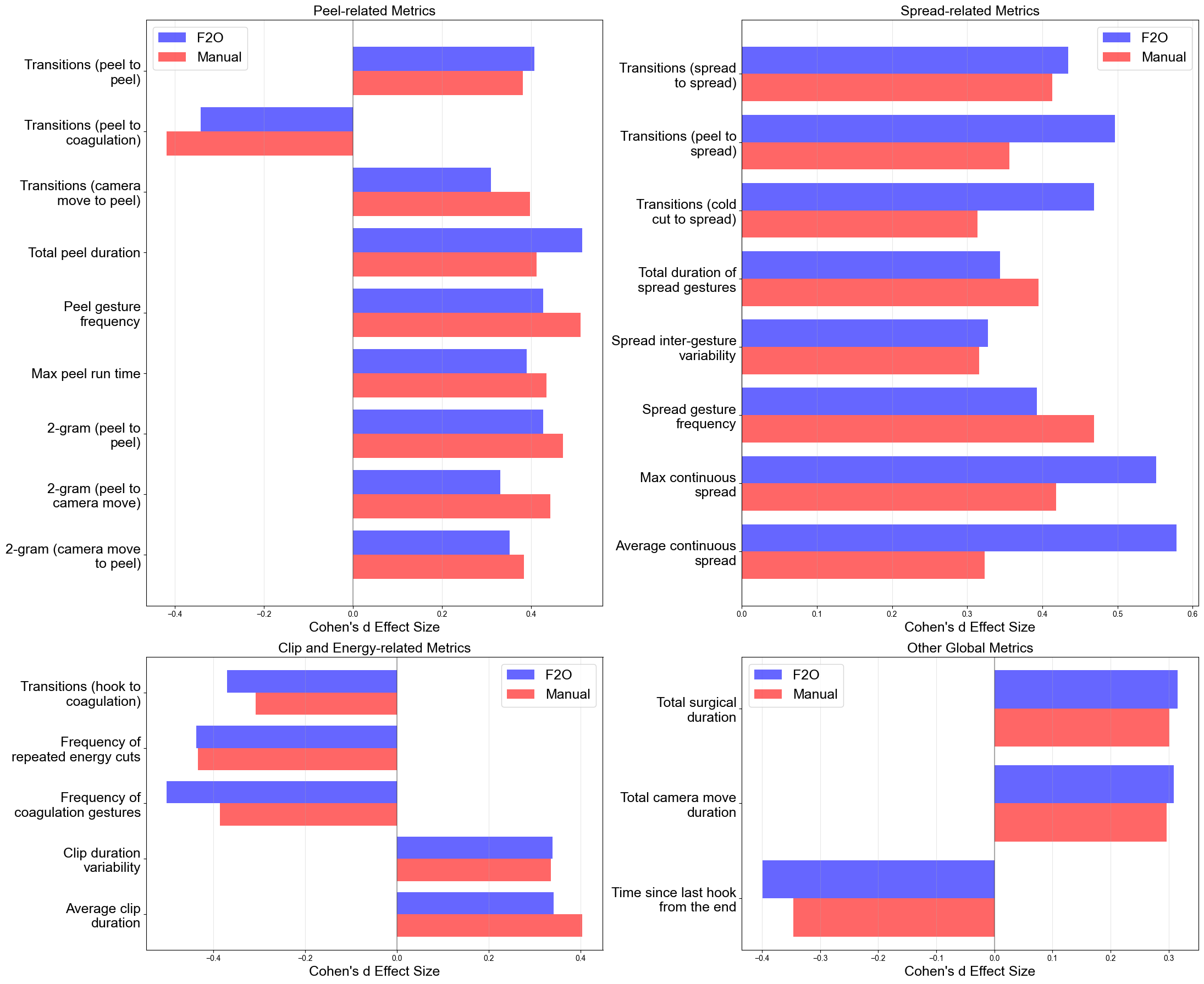}
    \caption{Effect size comparisons between the overlapping significant features of F2O and the ground truth gesture derived features. Positive effect size shows that the feature is correlated with better EF outcome. All features show the same direction and representation of populations.}
    \label{fig:cohend}
\end{figure}

Of the 50 features identified per set, 25 overlapped across comparisons between good and poor EF outcome groups (Fig. \ref{fig:cohend}), with high concordance based on effect size analysis. A positive effect size indicates that the feature has a higher value in the group with better EF outcomes. Table \ref{tab:overlap_features} and Fig. \ref{fig:cohend} categorize these overlapping features as follows:
\begin{enumerate}
\item \textit{Peel}-related metrics: Total \textit{peel} duration, frequency of \textit{peel} gestures, and maximum run time of \textit{peel} gestures are all global \textit{peel} metrics that are positively correlated with favorable EF outcomes. Two-gram features, which count any sequence of two consecutive gestures, indicate that transitions between \textit{peel} and \textit{camera move} in either direction are positively associated with good outcomes. Transition features, representing probabilities of moving from one gesture to another, show that \textit{peel} to \textit{peel} and \textit{camera move} to \textit{peel} transitions are positively correlated. In contrast, transitions from \textit{peel} to \textit{coagulation} are negatively correlated, likely reflecting that \textit{peel} gestures requiring coagulation, possibly due to bleeding, are associated with worse EF outcomes.
\item \textit{Spread}-related metrics: Average duration of continuous \textit{spread} gestures, maximum duration of continuous \textit{spread} gestures, total frequency of \textit{spread} gestures, and variability in the time between repeated \textit{spread} gestures all show positive correlation with favorable EF outcomes. Transition features such as \textit{spread} to \textit{spread}, \textit{peel} to \textit{spread}, and \textit{cold cut} to \textit{spread} are also positively correlated.
\item \textit{Clip}-related metrics: Greater variability in \textit{clip} duration and longer average \textit{clip} durations are observed in patients with favorable EF outcomes. This may indicate more deliberate and careful clipping.
\item Energy-related metrics: Higher frequency of \textit{coagulation} gestures, more repeated \textit{energy cuts}, and transitions from \textit{hook} to \textit{coagulation} are all negatively correlated with EF outcome. This suggests that increased use of energy-based tools may be associated with poorer outcomes.
\item Other global metrics: Longer total surgical duration and greater overall \textit{camera move} are positively correlated with favorable EF outcomes. 
\end{enumerate}

An analysis of the gesture types represented within these top 50 significant feature sets further underscored the systems' alignment. For the F2O features, those related to \textit{peel} gestures (14 features) and \textit{spread} (9 features) were most prevalent. Similarly, for the ground-truth features, \textit{peel} gestures (9 features) and \textit{spread} (7 features) were also the most frequently represented among the top significant features. This suggests that both systems converged on the importance of these particular gestures in relation to the clinical outcome.

To further evaluate the concordance between F2O and ground truth-derived features, we computed Cohen's d effect size for each of the 25 overlapping features, quantifying the magnitude of difference between outcome classes. These features represent the most clinically significant variables shared across both feature sets and are expected to convey a consistent narrative. A positive d value indicates a higher mean in the better outcome group compared to the poor outcome group. All effect directions (signs of d) were concordant between systems. The mean absolute difference in effect‐size magnitude was minimal (\(\Delta d_{\text{avg}} \approx 0.07\)). Cohen’s d values from Frame‑to‑Outcome and ground‑truth annotations were highly correlated (\(r = 0.96, n = 25, p < 1\times10^{-14})\), confirming that the AI‑derived features approximate manual annotations not only in statistical significance but also in clinical effect size (See Fig. \ref{fig:cohend}).

These 25 overlapping features reveal that deliberate, tissue-sparing gestures (longer \textit{peel}, \textit{spread}, and \textit{clip} actions; varied instrument sequences) favor erectile-function recovery, whereas frequent, repetitive energy use (\textit{coagulation} and \textit{energy cuts}) and hurried maneuvers increase the risk of postoperative dysfunction. F2O thus not only replicates manual scoring but pinpoints specific surgical behaviors with clear mechanistic ties to patient outcomes. This high degree of alignment further supports that F2O not only captures statistically significant features but also approximates their clinical relevance as quantified by effect size. These findings demonstrate that F2O reliably extracts clinically meaningful gesture features that closely match expert manual annotation, supporting its potential utility in automated clinical outcome prediction.

\begin{figure}[htbp]
\centering
\includegraphics[width=\textwidth]{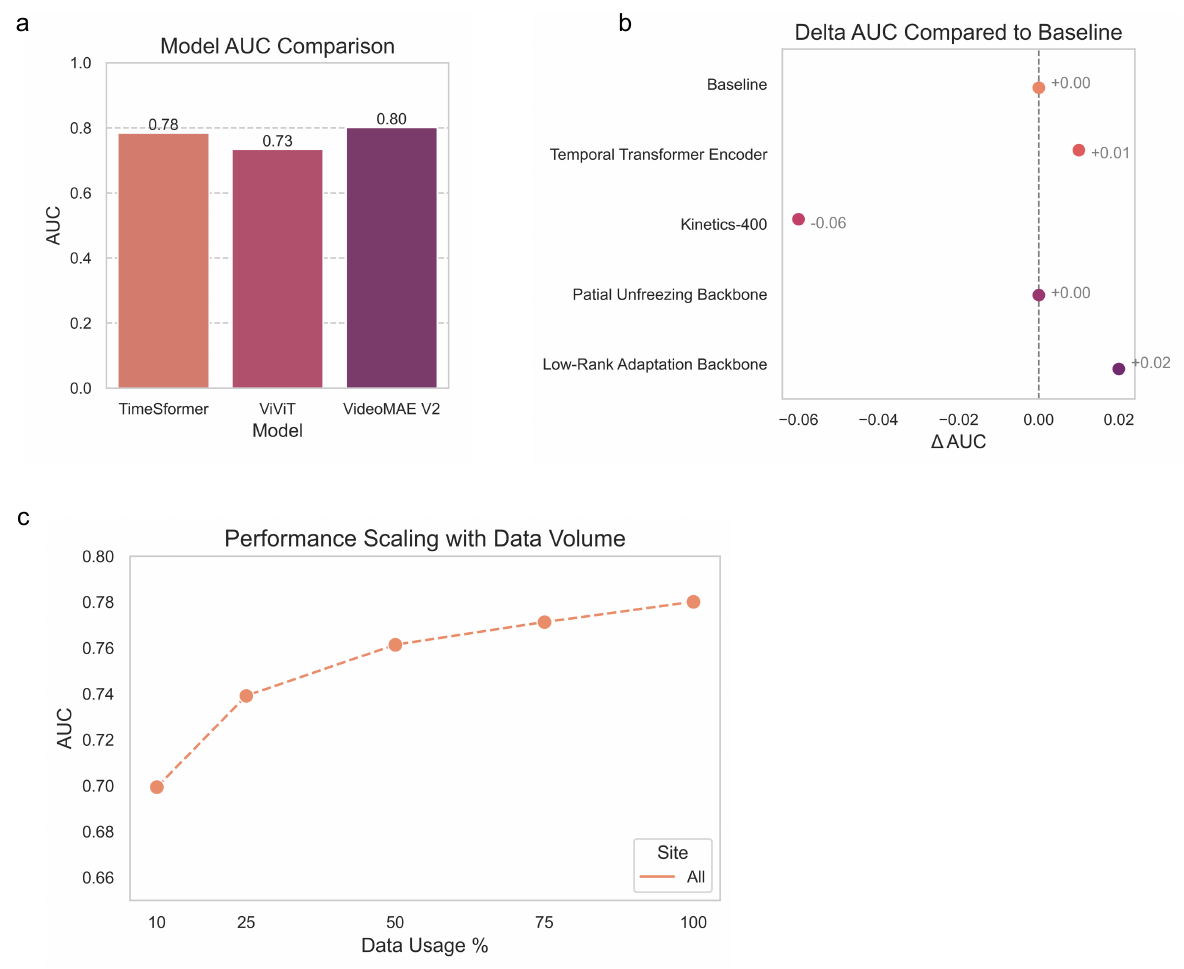}
\caption{Experiments across backbones and data scales showed consistent performance. \textbf{a}, Comparison of model performance across different transformer backbones, illustrating architectural flexibility. \textbf{b}, Impact of key design components on classification performance including the frame classifier architecture (Temporal Transformer Encoder), pretraining weights (Kinetics-400), and backbone optimization methods (Partial unfreezing, Low-Rank Adaptation), measured as relative changes from the baseline system. \textbf{c}, Model performance across increasing training data volumes, demonstrating robustness in low-data scenarios and scalability to new sites.
}\label{Fig. 4}
\end{figure}

\subsection{Generalizability across Transformer backbones}\label{subsec5}

To support long-term adaptability and computational efficiency, the F2O system was designed with a backbone-agnostic architecture capable of integrating emerging transformer models without retraining from scratch. 

Frame-level performance across three state-of-the-art backbones, VideoMAE V2 \cite{wang2023videomae}, TimeSformer \cite{bertasius2021space}, and ViViT \cite{arnab2021vivit}, is shown in Fig. \ref{Fig. 4}a. The system achieved AUCs of 0.79, 0.78, and 0.73, respectively. These results confirm the generalizability of the pipeline and its compatibility with evolving transformer architectures. This flexibility is particularly important in academic and clinical research environments, where access to large-scale computational infrastructure is often limited. For instance, pretraining a ViT-g model with VideoMAE V2 requires around two weeks on 64 GPUs. The ability to interchange backbones without modifying the rest of the system ensures long-term scalability and alignment with advances in video representation learning. 

\subsection{Impact of architectural and training modifications}\label{subsec6}

To quantify the contributions of key design components, we conducted studies (see Methods), measuring changes in frame-level classification AUC relative to the baseline configuration (Fig. \ref{Fig. 4}b). The components assessed included the frame classifier architecture, pretraining weights, and backbone optimization methods.

Incorporating a Temporal Transformer Encoder (TTE) and enhancing the backbone with Low-Rank Adaptation (LoRA) \cite{hu2022lora} both led to modest AUC gains, indicating that targeted improvements to temporal modeling and fine-tuning can further enhance classification accuracy. Conversely, using a backbone pretrained on Kinetics-400 \cite{kay2017kinetics} led to a 0.06 AUC drop compared to SSv2 \cite{goyal2017something}, highlighting the importance of domain-aligned pretraining. Unlike Kinetics-400’s coarse, long-duration actions, SSv2 contains fine-grained, short-term activities that better match the temporal characteristics of surgical gestures.

Collectively, these results demonstrate that F2O is both a high-performing and adaptable framework for surgical gesture recognition. By avoiding dependence on computationally intensive pretraining and supporting targeted adaptation techniques, the framework promotes sustainable model development and enhances reproducibility, key priorities as video representation learning continues to evolve.

\subsection{Scalability to sites with limited data}\label{subsec7}

A key translational challenge for AI in healthcare is deployment to sites with limited annotated data. To assess the model’s scalability under such constraints, we performed data scaling experiments using subsets of the training data: 10\%, 25\%, 50\%, 75\%, and 100\%. We evaluated performance with the AUC of the same test dataset (Fig. \ref{Fig. 4}c).

The model exhibited strong data efficiency. With only 10\% of the training data (\~24 cases), it achieved an AUC of 0.70 over 90\% of the full-data AUC (0.78). The largest performance gains occurred before reaching 50\% of the data. This behavior indicates that the system captures essential gesture patterns early in the learning curve, enabling rapid deployment even with limited training resources. These findings illustrate the model’s capacity to generalize to new clinical environments with minimal labeled data, facilitating broader adoption of surgical gesture recognition in diverse healthcare settings. 

\section{Discussion}\label{sec12}
An automatic method for capturing standardized, fine-grained actions across surgical videos lays the foundation for analyzing intraoperative behavior and its relationship to postoperative outcomes. While various AI-based approaches to analyze surgical video have been explored, a systematic, standardized method to link intraoperative video to outcomes remains elusive due to persistent challenges in terminology, annotation, data variability, and modeling complexity \cite{ward2021challenges,loukas2018video,nyangoh2023systematic}. These gaps have collectively delayed the integration of intraoperative activity-based analysis into outcome research, limiting the ability to understand how specific surgical actions influence recovery, complications, and overall patient outcomes.

This study introduces F2O, an end-to-end system that translates intraoperative video into a sequence of standardized surgical gesture classes (average duration: 2 seconds), providing interpretable and clinically meaningful signals embedded in surgical behavior. This system addresses the longstanding need for automated, standardized intraoperative measurements and their relationship to clinical outcomes \cite{balvardi2022association}. The results have demonstrated that F2O 1) achieves accurate frame-level classification across surgical gesture classes, 2) enables automatic gesture sequence recognition across videos, and 3) supports interpretable downstream analysis of surgical outcomes. 

In contrast to prior methods reliant on coarse time resolution and manually annotated boundaries \cite{loukas2020surgical,hashimoto2019computer,markarian2022validation}, F2O leverages vision transformer-based spatial and temporal modeling and frame-wise classification to provide high-resolution predictions with mean AUC 0.80 (95\% CI: 0.78-0.81) across ten dominant surgical gestures, which account for over 97\% of gestures in the dataset.  When applied to full-length tissue dissection videos, the system produces smooth and temporally coherent predictions that align with ground truth annotations (Fig. \ref{Fig. 2}b), achieving strong performance with a median AUC of 0.81 across 29 RARP NS videos, despite heterogeneity in case complexity and gesture distribution. This automated process effectively replaces the need for manual gesture annotation in surgical video analysis which requires the development of procedure-specific guides for training personnel \cite{pangal2021guide,fischer2023methodology}. 

Beyond classification accuracy, our results demonstrate that F2O not only replicates the statistical significance of manual annotations (high overlap in top-ranked features and r = 0.96 Pearson correlation of Cohen’s d values) but also uncovers the same clinically relevant behaviors that experienced surgeons intuitively value. In particular, deliberate, sustained gestures—longer cumulative \textit{peel} durations (\texttt{dur\_sum\_p}, d = –0.51 vs. –0.41 (F2O vs Ground truth respectively)) and extended \textit{spread} runs (\texttt{avg\_run\_s}, d = –0.58 vs. –0.32) were enriched in the favorable-outcome cohort. These metrics align with careful tissue manipulation around neurovascular bundles and support nerve preservation \cite{quinlan1991sexual, steineck2015degree,michl2016nerve}. Conversely, frequent \textit{coagulation} gestures (\texttt{freq\_g}, d = +0.50 vs. +0.39) and repeated energy transitions (\texttt{trans\_e\_e}, d = +0.44 vs. +0.43) correlated with poorer functional recovery, likely reflecting energy related damage to the nerves \cite{ahlering2006impact}. Thus, F2O indicate similar interpretable behavioral markers to the ground truth dataset, such as transition ratios and n-gram patterns, that map directly onto surgical technique and patient outcome.

From a clinical perspective, the most intractable challenge has been the delayed nature of surgical end-points, which often only manifest months or years postoperatively \cite{sandhu2010factors,ko2012factors}. Traditional feedback to the surgical team is therefore episodic and anecdotal, lacking granularity on which segment or instrument maneuver drove success or complication. By contrast, our sequential-analysis approach digitizes every frame into gesture events and quantifies their impact on the eventual clinical outcome. This creates a closed feedback loop: surgeons can review which specific \textit{peel} or \textit{spread} gestures differentiated high-functioning cases, and they can iteratively refine their technique in near real time. Such granular and automated feedback has the potential to accelerate skill acquisition beyond what retrospective chart review can achieve \cite{wang2020towards,chen2024decoding}.

In addition to its clinical relevance, F2O’s modular architecture underpins its extensibility. The frame-to-gesture model, the change-point detection, and gesture-based feature engineering can be retrained or altered on diverse endoscopic procedures from laparoscopy, bronchoscopy, to arthroscopy with minimal code changes, supported by a predefined standardized dissection gesture taxonomy. Outcome labels are equally flexible, extending beyond erectile functional recovery metrics to include hospital length of stay after surgery, post-operative complication rates, and patient-reported outcomes \cite{krell2014extended}. Consistent performance across various backbones demonstrates the system’s plug-in capability, eliminating the need for weeks of computation on high-end GPUs \cite{yang2024effictran}. As more pre-trained vision models and larger cohorts become available, the pipeline can be updated in a plug-and-play fashion, ensuring continuous improvement without imposing new burdens on the user \cite{modi2023towards} and supporting resource-limited clinical environments. Furthermore, F2O exhibits strong data efficiency. With only 10\% of training data (\~24 videos), it achieved over 90\% of the full-data AUC. This enables practical implementation in new clinical sites where large-scale data collection is not feasible.

Despite these strengths, several limitations warrant discussion. First, our cohort (N = 138) focused on a single procedure type and a binary outcome measure; larger, multi-center studies are needed to validate generalizability across institutions and more nuanced clinical endpoints. Second, although we achieved high agreement with manual annotation, black swan events, rare, or subtle gestures may escape robust detection due to limits on the number of types of gestures able to be analyzed and limited data to train the model on such gestures. Future algorithms should integrate temporal context or unsupervised gesture discovery to capture these events. Third, real-time deployment poses engineering challenges (latency, hardware integration) and must be paired with user‐interface design that ensures surgeon buy-in and seamless workflow integration. Finally, clinical deployment may be limited by technical and computational barriers. A user-friendly interface that allows secure, de-identified inference and analysis of surgical videos would be particularly beneficial, especially for early-career surgeons. 

Looking forward, we can think of some examples where F2O's framework show promise:

\begin{enumerate}
    \item Prospective Validation and Clinical Trials. Embedding F2O into live cases and correlating its real‐time feedback with subsequent outcomes will establish its efficacy as a training aid and quality‐assurance tool.

    \item Adaptive, Outcome-Weighted Gesture Coaching. By linking specific gesture scores to predictive risk, one could create a dashboard that alerts the surgeon if their current sequence deviates from patterns associated with good outcomes, enabling immediate corrective action.

    \item Cross-Domain Transfer and Meta-Analysis. Aggregating gesture‐outcome mappings across multiple procedure types may reveal universal principles of tissue handling and instrument use, informing best-practice guidelines and AI-augmented surgical curriculum.
    
    \item Public Tool Integration for Training and Decision Support. Deploying F2O within publicly accessible platforms could support surgical training, quality assurance, and outcome-informed decision-making by providing interpretable feedback and scalable gesture analytics across diverse clinical settings. 
\end{enumerate}

In summary, F2O bridges the gap between endoscopic surgery and the delayed nature of clinical endpoints. By quantifying and interpreting every surgical gesture, it offers an unprecedented window into the subtle, yet clinically critical, elements of operative technique that drive clinical outcomes. As AI models continue to evolve, such hands-off, data-driven systems will become indispensable for surgical education, quality control, and ultimately, patient care.

\section{Methods}\label{sec11}

\subsection{Surgical gesture data}\label{subsec2}

Surgical motion revolves around moment-to-moment gestures, which serve as the fundamental units of action, similar to binary code in computing. Decoding procedures at the gesture level enables standardized representations of surgical workflows and enhances both performance evaluation and postoperative outcome prediction.

Videos from the nerve-sparing (NS) section of robot-assisted radical prostatectomy (RARP) exhibit rich gesture-level activity, with approximately 270 gestures in a 10-minute video, each averaging 2 seconds (Fig. \ref{Fig. 1}a). These gestures flow continuously and, when annotated, provide valuable insights for outcome prediction and performance assessment. 

Previous work established a dissection surgical gesture classification system consisting of dissection gestures (e.g., \textit{cold cut}) and supporting gestures (e.g., \textit{retraction}) (Fig. \ref{Fig. 1}a)  and demonstrated the relationship between gestures used during the NS step of RARP and postoperative erectile dysfunction rates.

In this study, we used 294 annotated RARP-NS videos collected from four international institutions (University of Southern California [USC], Memorial Sloan Kettering Cancer Center [MSK], St. Antonius Hospital [SAH], and Stanford University Medical Center [SMC]). The dataset includes surgeries performed by 23 unique surgeons, representing the diverse real-world scenario. Ground-truth gesture annotations were provided by trained human raters, who labeled the start and end times of each gesture following strict guidelines defined in our previously developed taxonomy of dissection gestures. On average, each video contains \~270 gestures, each lasting about 2 seconds. Over 97\% of gestures belong to 10 dominant classes, including \textit{cold cut} (c), \textit{hook} (h), \textit{clip} (k), \textit{camera move} (m), \textit{peel} (p), \textit{retraction} (r), \textit{spread} (s), \textit{assistant} (a), \textit{coagulation} (g), and \textit{energy cut} (e). The remaining 3\% are labeled as \textit{X} and excluded from automated gesture recognition due to insufficient training data for these infrequent classes.

\subsection{Spatial and temporal modeling}\label{subsec2}

To capture high-level spatial features and temporal dependencies between frames, the system employs spatial and temporal modeling using vision Transformers, which interpret video as a sequence of patches extracted from the individual frames, and each patch is mapped into an embedding and augmented with positional information. 

We adopted TimeSformer\cite{bertasius2021space}, a vision transformer specifically designed for general video understanding. Given a sequence of \( F \) RGB frames of size \( H \times W \), TimeSformer divides each frame into \( N = \frac{H \times W}{P^2} \) non-overlapping patches of size \(P \times P\). Each Patch \( \mathbf{x}_{(p,t)} \in \mathbb{R}^{3P^2} \) is linearly projected into an \( D \)-dimensional embedding:
\[
\mathbf{z}^{(0)}_{(p,t)} = \mathbf{E} \mathbf{x}_{(p,t)} + \mathbf{e}^{\text{pos}}_{(p,t)}
\]
where \( \mathbf{E} \in \mathbb{R}^{D \times 3P^2} \) a learnable matrix, \( \mathbf{e}^{\text{pos}}_{(p,t)} \in \mathbb{R}^D \) is a learnable spatiotemporal positional embedding, \( p \in \{1, \dots, N\} \) is the spatial patch index, and \( t \in \{1, \dots, F\} \) is the frame index.

Temporal and spatial attention modules are applied sequentially, enabling the model to learn rich spatiotemporal features from frame sequences. An example visualization is shown in Fig. \ref{Fig. 1}b, spatial and temporal modeling, where red and green highlighted patches are used for self-attention computation of the blue patch. After processing through \( L \) layers of divided space-time attention, the Transformer outputs a sequence for batch size \( B \):

\[
\mathbf{Z} = \left[ \mathbf{z}^{(L)}_{(0,0)},\ \mathbf{z}^{(L)}_{(1,1)},\ \dots,\ \mathbf{z}^{(L)}_{(N,F)} \right] \in \mathbb{R}^{B \times (NF + 1) \times D}
\]

We selected \textbf{TimeSformer-HR}, a high resolution variant that operates on 16 frames of size 448×448, over default \textbf{TimeSformer} (8 frames at 224×224) and the long range \textbf{TimeSformer-L} (96 frames at 224×224). TimeSformer-HR strikes a balance between temporal coverage and spatial fidelity, allowing it to model longer temporal dependencies without being distracted by minor inter-frame variations. For pretraining, we selected Something-Something V2 (SSv2) as pretrain weights, as its temporal complexity more closely resembles the high-frequency and short-duration surgical activity, compared to Kinetics-400 (K400), known for capturing general human actions. 

With the selected pre-trained backbone, our system processes each frame to extract high-level spatial features. Temporal relationships between frames are captured to understand action evolution over time. This dual approach ensures that both spatial and temporal information contribute to action recognition.

\subsection{Frame-wise classification}\label{subsec2}

Frame-wise classification plays a key role in detecting transitions between actions, especially given that gestures average only 2 seconds and often change rapidly. For instance, a \textit{clip} gesture may transition to a cut gesture in less than a second, necessitating high temporal resolution in classification. Previous efforts have attempted to learn frame-level representations, but surgical activity recognition at this granularity has not yet been extensively developed.

In our framework, temporal features extracted from TimeSformer are aggregated per frame. We excluded the classification token \( \mathbf{z}^{(L)}_{(0,0)} \) and reshaped the remaining output tokens to recover frame-wise structure:
\[
\mathbf{Z}' = \text{reshape}(\mathbf{Z}_{[:,1:,:]},\ (B,\ F,\ N,\ D))
\]
To obtain frame-level features, patch embeddings are averaged within each frame:
\[
\mathbf{F}_t = \frac{1}{N} \sum_{p=1}^{N} \mathbf{Z}'_{[:, t, p, :]} \quad \text{for } t = 1, \dots, F
\]
This produces the final frame-level representation:
\[
\mathbf{F} \in \mathbb{R}^{B \times F \times D}
\]

To classify gesture classes at the frame level, we explored two lightweight frame-wise classifiers. 
\begin{itemize}
    \item A simple linear layer each 768-dimensional feature vector to a 10-dimensional vector of logits, effectively learning to classify each frame into one of the 10 gesture categories.  
    \item A lightweight Temporal Transformer Encoder (TTE) on top of the TimeSformer backbone consists of a single Transformer encoder layer with four attention heads to capture diverse temporal patterns across gesture sequences. This is followed by a two-layer multilayer perceptron (MLP) that maps the temporally encoded frame features to class probabilities. The MLP includes a 256-dimensional hidden layer with ReLU activation and dropout regularization (p = 0.3), followed by a final linear projection to the number of gesture classes.
\end{itemize}

These lightweight classifiers maintain computational efficiency while maximizing the utility of the pretrained backbone for fine-grained frame-level recognition.

\subsection{Video dataset processing and frame classifier training pipeline}\label{subsec2}

The videos were split into training (80\%), validation (10\%), and test (10\%) sets. Each frame was labelled according to the manually annotated start and end times. Stratified and adaptive overlapping sequence sampling was performed at adjusted FPS, adjusted sampling stride based on class weights to handle diverse FPS across different hospitals, to enrich the minority classes, and to ensure an equal number of cases per gesture class. This step is to ensure diversity and sufficient data coverage across the train, validation, and test datasets, for example, the dataset comprising 80,000 frames per class in the training dataset for the 10 most common gestures, validation and test sets containing ~8,000 frames per class. Data augmentation techniques, including zoom and rotation, were applied to improve generalizability.

Model training was conducted using a distributed data parallel (DDP) framework with mixed precision (AMP) to enable efficient multi-GPU training. We optimized a class-balanced cross-entropy loss using the Adam optimizer with weight decay and a cosine annealing learning rate schedule with linear warm-up. During training, data were shuffled to ensure representative sampling. Performance was evaluated on a validation set using frame-level area under the ROC curve (AUC) and loss. Early stopping was applied based on validation loss, and model checkpoints were saved every five epochs. In addition to AUC, we computed per-class AUC on test dataset to assess class-specific performance (Fig. \ref{Fig. 2}a). 

\subsection{Automatic gesture sequence recognition in a video}\label{subsec2}

Once the frame-level classification model was trained, it was applied to full-length, untrimmed videos using a sliding window approach. Videos were uniformly sampled at a fixed time interval (e.g., 0.1667 s), and non-overlapping frame sequences of length 16 were passed through the model.  The model processes each sequence by capturing both spatial context and motion patterns, aggregating temporal embeddings, and producing frame-by-frame gesture class probabilities. For each video, class-wise AUCs were computed only for gesture classes present in the ground truth. The mean AUC across these present classes was then reported as the video-level AUC (Fig. \ref{Fig. 2}b). This approach ensures that model performance is assessed fairly and without penalization for absent classes. For each video, we generated visualizations of predicted gesture probability trajectories and exported structured probability files, which serve as inputs for downstream tasks such as change point detection and clinical outcome prediction. 

\subsection{Comparison across backbones}\label{subsec2}

The system leverages pretrained foundational models to maintain flexibility in adopting state-of-the-art architectures while minimizing computational demands. Recent vision Transformer models are typically trained on substantial hardware resources. For instance, TimeSformer, published in 2021, was trained using 32 GPUs, while pretraining a ViT-g model with VideoMAE V2 (2023) requires around two weeks on 64 A100 GPUs. In contrast, academic institutions often rely on limited and aging GPU infrastructure, creating barriers to adopting such resource-intensive workflows.
To address this, our system is designed to maximize the utility of publicly available pretrained models. We benchmarked three Transformer backbones(TimeSformer, ViViT, and VideoMAE) within our frame-wise classification pipeline. To ensure a fair comparison, each backbone was frozen during training, and only the lightweight linear frame classifier was trained. Training and evaluation were conducted using consistent hyperparameters and data processing settings. Results are reported as frame-level AUCs, visualized in a boxplot (Fig. \ref{Fig. 4}a).

\subsubsection{VideoMAE backbone}\label{subsubsec2}

To evaluate the generalizability of our framework, we implemented a version of the model using VideoMAE-v2 as the backbone for frame-wise gesture classification. We adapted the vit\_giant\_patch14\_224 model from the official VideoMAE repository and initialized it with pretrained weights fine-tuned on SSv2. The model processes sequences of 16 video frames and extracts spatiotemporal patch embeddings via the transformer encoder. These embeddings are pooled per frame to form a compact temporal representation, which is then passed through a lightweight linear classifier to predict gesture labels. 

\subsubsection{ViViT backbone}\label{subsubsec3}

We also implemented a model variant using ViViT as the backbone for frame-wise gesture classification. We leveraged a pretrained ViViT model from Hugging Face to extract spatiotemporal patch embeddings from sequences of 16 video frames. The tokenized output, excluding the \(CLS\) token, was reshaped to recover per-frame representations by averaging over patches. These frame-level features were then passed through a lightweight frame classifier to classify each frame into gesture categories. This implementation allowed us to compare ViViT with other backbones under a consistent architecture and training setup.

\subsection{Model variant studies}\label{subsec2}

To evaluate the contribution of individual components to overall performance, we conducted a series of experiments around key components frame classifiers, pre-trained weights, and backbone optimization. The experiments included adding temporal transformer encoder in the frame classifier, comparing different pretrained weights (TimeSformer fine-tuned on SSv2 vs. Kinetics-400), and exploring various backbone fine-tuning strategies such as partial unfreezing and Low-Rank Adaptation (LoRA). We used TimeSformer-HR SSv2 with a linear frame classifier as the baseline, keeping similar hyperparameters, data processing, and augmentation.  These experiments provided insights into how each architectural and training decision influenced temporal modeling and gesture classification performance. 

\subsection{Data scaling}\label{subsec2}

Data scaling experiments were conducted to assess how model performance varies with increasing amounts of training data. Our dataset comprises 294 NS sections of RARP surgeries collected from four international sites, involving approximately 23 surgeons. The number of cases per site and per surgeon varies substantially from around 10 to 140 surgeries, due to the complexity and difficulty of surgical video data collection and cross-institutional sharing.

To evaluate the model’s scalability to new sites with limited data, we assessed its performance under restricted training data conditions, simulating deployment in settings where only small-scale datasets are available. We performed stratified random splits at the video level to divide the dataset into train/validation/test sets (80/10/10), ensuring balanced representation across sites. The test set was fixed across all experiments to enable fair comparisons between different data scales.

The model was trained using 10\%, 25\%, 50\%, 75\%, and 100\% of the training set, while keeping validation and test sets unchanged. For all experiments, we used the baseline TimeSformer-HR pretrained on SSv2, with a linear frame-level classifier, and maintained consistent hyperparameters, data processing steps, and augmentation strategies. Model performance was evaluated using AUC on the test set.

\subsection{Gesture aggregation via change point detection}\label{subsec2}
In order to train downstream applications such as clinical outcome prediction, frame-level gesture predictions need to be aggregated into temporally meaningful gesture sequences akin to human-annotated gesture sequences. To create aggregated gesture sequences, where each gesture spans 2-3 seconds, the model's class probability distributions associated with each frame were leveraged.
Through change point detection, significant shifts in the probability distribution from one gesture to another were identified and used to group frames into dominant gestures. 

An implementation of linearly penalized segmentation (Pelt) in the ruptures python library was used due to its ability to identify shifts within non-linear distributions \cite{TRUONG2020107299, Killick01122012}. The Pelt model parameter was set to \textit{rbf}. The penalty value, which determines the sensitivity of the change point detector, was fixed after empirically assessing values in the range [0,1]. A penalty value of 0.5 resulted in a reasonable alignment between the lengths of the aggregated and human-annotated gesture sequences.  

\subsection{Feature engineering for clinical outcome prediction}\label{subsec2}
To capture the temporal and sequential dynamics of user gestures, we engineered a comprehensive set of features from raw gesture, timestamp, and duration data. These features were categorized into several domains to reflect different behavioral and statistical properties of gesture sequences (Fig. \ref{Fig. 1}c). In total, 2484 features were created per case.

\paragraph{Frequency-Based Features}

We computed normalized frequencies for each gesture type to quantify their relative occurrence over the entire sequence. This included global gesture frequencies, as well as exponential decay-weighted counts that emphasize more recent gestures. The decay-weighted approach ensures temporal sensitivity by assigning higher importance to gestures closer to the present moment.

\paragraph{Temporal Features}

Time-based features captured both global and gesture-specific timing information. These included:

\begin{itemize}
    \item Total sequence duration - time between the first and last gesture
    \item Gesture rate - gestures per unit time
    \item Time since last occurrence for each gesture, calculated relative to the current timestamp.
\end{itemize}
These features characterize user activity intensity and temporal recency of individual gestures.

\paragraph{Sequence Structure Features}

To quantify the structural properties of gesture sequences, we included:

\begin{itemize}
    \item The number of unique gestures used
    \item The total number of gesture changes
    \item The Shannon entropy of the gesture distribution.
\end{itemize}
These metrics reflect both the diversity and predictability of the user's gestural behavior.

\paragraph{N-gram and Transition Features}

We incorporated n-gram statistics (for n=2 and 3) to capture local sequential dependencies. Each n-gram feature represents the normalized frequency of a contiguous subsequence of gestures. Transition probabilities between gesture pairs were estimated, modeling the likelihood of one gesture following another. These features reflect the short-term temporal dynamics of gesture transitions.

\paragraph{Dwell-Time Features}

We computed inter-gesture timing statistics (dwell times) to capture rhythm and pacing:

\begin{itemize}
    \item Mean, standard deviation, minimum, maximum, and median of dwell times
    \item Higher-order moments including skewness and kurtosis of dwell distributions
    \item Gesture-specific dwell time statistics preceding each gesture type
\end{itemize}
These features describe the temporal granularity of user interactions and may indicate hesitation or fluidity in motion.

\paragraph{Gesture Duration Features}

We extracted aggregate statistics for the duration of each gesture, including mean, standard deviation, skewness, kurtosis, and total duration. These features were computed both globally and separately for each gesture class, enabling characterization of gesture-specific temporal patterns and variability.

\paragraph{Run-Length Encoding Features}

To model gesture repetition and persistence, we computed run-length features per gesture:

\begin{itemize}
    \item The maximum run length 
    \item The average run length of consecutive occurrences
\end{itemize}
These features quantify how often users repeat or sustain gestures, which may relate to intent or error states.

\section*{Declarations}
\begin{itemize}
\item \textbf{Proprietary notice} \\
This end-to-end AI system for surgical gesture sequence recognition and clinical outcome prediction, known as Frame-to-Outcome (F2O), is owned by and proprietary to Cedars-Sinai Medical Center. © 2025 Cedars-Sinai Medical Center. All rights reserved.
For any use requests, please reach out to CSTechTransfer@cshs.org
\\
\item \textbf{Funding} \\
Research reported in this publication was supported by the National Cancer Institute of the National Institutes of Health under Award Number R01CA273031. The content is solely the responsibility of the authors and does not necessarily represent the official views of the National Institutes of Health.
\\
\item \textbf{Competing interests} \\
All authors declare no financial or non-financial competing interests. 
\\
\item \textbf{Ethics approval and consent to participate} \\
Not applicable
\\
\item \textbf{Consent for publication} \\
All authors reviewed and approved the final manuscript.
\\
\item \textbf{Data availability} \\
The datasets generated and analyzed during the current study which contain patient information are not publicly available. De-identified data may be made available upon reasonable request to the corresponding author.
\\
\item \textbf{Materials availability} \\
Not applicable
\\
\item \textbf{Code availability} \\
The underlying code for this study is not publicly available but may be made available to qualified researchers on reasonable request from the corresponding author.
\\
\item \textbf{Author contribution} \\
A.J.H. conceived the study, provided supervision, and contributed to manuscript revisions. X.L. and N.M. contributed to the study conception, design, model development, and manuscript writing. U.P. contributed to change-point detection and manuscript writing. A.D. contributed to model evaluation. J.M. contributed to LoRA component. C.Y. and J.K. contributed to data annotation and management. M.E.H. and P.W. provided project oversight and resource management. J.L. provided feedback on the manuscript. A.C.G., C.W., and G.A.S. provided data for the study. 
\end{itemize}

\begin{appendices}

\section{Comparison of overlapping features between AI system and Ground Truth Based on Recovery of Erectile Function in 12 months (EF)}\label{secA2}

\begin{sidewaystable}
\caption{Comparison of overlapping features between AI system and Ground Truth Based on Recovery of Erectile Function in 12 months (EF)}\label{tab:overlap_features}
\begin{tabular*}{\textheight}{@{\extracolsep\fill}lcccccc}
\toprule
& \multicolumn{3}{@{}c@{}}{F2O} & \multicolumn{3}{@{}c@{}}{Ground Truth} \\\cmidrule{2-4}\cmidrule{5-7}
Feature & Poor EF & Good EF & $p$-value & Poor EF & Good EF & $p$-value \\
\midrule
dur\_sum\_p         & 142.020 ± 138.948 & 243.067 ± 240.817 & 0.0023 & 202.064 ± 181.159 & 292.157 ± 250.448 & 0.0187 \\
avg\_run\_s         &   0.934 ± 0.640   &   1.274 ± 0.526   & 0.0031 &   1.181 ± 0.562   &   1.352 ± 0.496   & 0.0902 \\
max\_run\_s         &   1.542 ± 1.704   &   2.476 ± 1.685   & 0.0035 &   2.375 ± 2.341   &   3.333 ± 2.238   & 0.0266 \\
trans\_p\_s         &   0.0115 ± 0.0271 &   0.0293 ± 0.0427 & 0.0038 &   0.0197 ± 0.0229 &   0.0299 ± 0.0332 & 0.0398 \\
trans\_c\_s         &   0.0037 ± 0.0143 &   0.0118 ± 0.0200 & 0.0075 &   0.0152 ± 0.0250 &   0.0240 ± 0.0308 & 0.0791 \\
2gram\_p\_p         &   0.0636 ± 0.0602 &   0.0993 ± 0.1021 & 0.0113 &   0.1594 ± 0.0852 &   0.1993 ± 0.0842 & 0.0122 \\
freq\_g             &   0.1557 ± 0.1213 &   0.1047 ± 0.0771 & 0.0133 &   0.0419 ± 0.0375 &   0.0292 ± 0.0279 & 0.0499 \\
freq\_p             &   0.1451 ± 0.0924 &   0.1926 ± 0.1277 & 0.0150 &   0.2954 ± 0.1049 &   0.3504 ± 0.1104 & 0.0060 \\
trans\_s\_s         &   0.1263 ± 0.1764 &   0.2052 ± 0.1868 & 0.0190 &   0.1379 ± 0.1895 &   0.2161 ± 0.1893 & 0.0273 \\
trans\_e\_e         &   0.1616 ± 0.2136 &   0.0775 ± 0.1687 & 0.0254 &   0.1199 ± 0.1980 &   0.0495 ± 0.1167 & 0.0335 \\
max\_run\_p         &   5.052 ± 3.689   &   6.810 ± 5.214   & 0.0255 &   8.594 ± 4.383   &  10.571 ± 4.733   & 0.0187 \\
trans\_p\_p         &   0.3435 ± 0.1786 &   0.4165 ± 0.1800 & 0.0293 &   0.5069 ± 0.1145 &   0.5473 ± 0.0971 & 0.0483 \\
freq\_s             &   0.0181 ± 0.0240 &   0.0278 ± 0.0255 & 0.0333 &   0.0343 ± 0.0328 &   0.0513 ± 0.0393 & 0.0096 \\
time\_since\_last\_h&  30.713 ± 355.326 & –94.849 ± 267.687 & 0.0425 & –12.554 ± 329.848 & –129.851 ± 348.545 & 0.0610 \\
dur\_std\_k         &   1.2926 ± 0.8803 &   1.7014 ± 1.4653 & 0.0446 &   3.5122 ± 2.8926 &   4.6971 ± 4.0899 & 0.0543 \\
dur\_mean\_k        &   2.0772 ± 0.7251 &   2.3796 ± 1.0266 & 0.0503 &   7.8145 ± 4.7732 &  10.6927 ± 8.9094 & 0.0150 \\
trans\_h\_k         &   0.1460 ± 0.0947 &   0.1138 ± 0.0784 & 0.0552 &   0.1563 ± 0.1611 &   0.1116 ± 0.1268 & 0.1138 \\
2gram\_m\_p         &   0.0349 ± 0.0263 &   0.0446 ± 0.0287 & 0.0555 &   0.0217 ± 0.0116 &   0.0270 ± 0.0156 & 0.0291 \\
dur\_sum\_s         &  14.3056 ± 30.7969 &  25.4443 ± 33.9446 & 0.0603 &  34.6293 ± 45.0479 &  52.1424 ± 43.6945 & 0.0358 \\
2gram\_p\_m         &   0.0356 ± 0.0266 &   0.0445 ± 0.0269 & 0.0754 &   0.0371 ± 0.0148 &   0.0436 ± 0.0145 & 0.0184 \\
dwell\_before\_std\_s&  78.2188 ± 145.3837 & 124.9827 ± 139.9487 & 0.0810 &  75.3926 ± 103.7871 & 120.2381 ± 171.9135 & 0.0608 \\
duration\_sum       & 846.0862 ± 590.8864 & 1033.8804 ± 598.6577 & 0.0893 & 952.2324 ± 648.7883 & 1162.9625 ± 748.5010 & 0.0964 \\
trans\_p\_g         &   0.1460 ± 0.1579 &   0.1010 ± 0.0990 & 0.0907 &   0.0361 ± 0.0489 &   0.0197 ± 0.0262 & 0.0429 \\
dur\_sum\_m         & 251.6336 ± 232.3093 & 324.8925 ± 241.8451 & 0.0946 & 115.4164 ± 77.2925 & 138.2658 ± 76.8012 & 0.1117 \\
trans\_m\_p         &   0.1246 ± 0.0832 &   0.1507 ± 0.0847 & 0.0950 &   0.1588 ± 0.0761 &   0.1936 ± 0.0978 & 0.0253 \\
\botrule
\end{tabular*}
\footnotetext{Values reported as mean ± standard deviation}
\end{sidewaystable}




\end{appendices}


\clearpage

\bibliography{sn-bibliography}

\end{document}